**E. G. Abramov[1*], A. B. Komissarov[2], D. A. Kornyakov**

# Generalized version of the support vector machine for binary classification problems: supporting hyperplane machine.

In this paper there is proposed a generalized version of the SVM for binary classification problems in the case of using an arbitrary transformation $\mathbf{x} \longmapsto \mathbf{y}$. An approach similar to the classic SVM method is used. The problem is widely explained. Various formulations of primal and dual problems are proposed. For one of the most important cases the formulae are derived in detail. A simple computational example is demonstrated. The algorithm and its implementation is presented in Octave language.

[1] Laboratory of General Biophysics, Department of Biophysics, Faculty of Biology, Saint-Petersburg State University, Saint-Petersburg, Russia.

[2] Laboratory of Molecular Virology and Genetic Engineering, Research Institute of Influenza, Saint-Petersburg, Russia.

[*]Corresponding author e-mail: evgeniy.g.abramov@gmail.com

## Contents.

## 1. Introduction.

The Support Vector Machine (SVM) which was proposed by Cortes and Vapnik [1] is nowadays one of the most powerful tools to artificial neural networks. Its simplicity and efficiency attract many researchers and engineers in various sciences and technologies.

Based on the initial approach, we propose a generalized version of the SVM classic binary classification problem of the following type.

---

If the training set $\left\{(\mathbf{x}_i, d_i)\middle|\mathbf{x}_i \in \mathbb{R}^m, d_i \in \{-1,1\}\right\}_{i=1}^{N}$ is known find out the hyperplane family

$$\mathbf{x}^{\mathrm{T}}\mathbf{W}\mathbf{f}(\mathbf{x}) + \mathbf{w}_0^{\mathrm{T}}\mathbf{x} + b = 0, \tag{1.1}$$

under the constraints

$$\begin{aligned} \mathbf{x}_{\mathrm{i}}^{\mathrm{T}}\mathbf{W}\mathbf{f}(\mathbf{x}_{\mathrm{i}}) + \mathbf{w}_0^{\mathrm{T}}\mathbf{x}_{\mathrm{i}} + b \geq 1 \text{ for } d_{\mathrm{i}} = +1 \\ \mathbf{x}_{\mathrm{i}}^{\mathrm{T}}\mathbf{W}\mathbf{f}(\mathbf{x}_{\mathrm{i}}) + \mathbf{w}_0^{\mathrm{T}}\mathbf{x}_{\mathrm{i}} + b \leq -1 \text{ for } d_{\mathrm{i}} = -1 \end{aligned} \tag{1.2}$$

and minimize the functional

$$\Phi(\mathbf{W}, \mathbf{w}_0) = \frac{1}{2}\sum_{i=1}^{N}\left[(\mathbf{x}_i^{\mathrm{T}}\mathbf{W})(\mathbf{x}_i^{\mathrm{T}}\mathbf{W})^{\mathrm{T}} + (\mathbf{w}_0^{\mathrm{T}}\mathbf{x}_i)^2\right] \tag{1.3}$$

under the condition that the function set $\mathbf{y} = \mathbf{f}(\mathbf{x})$ is known.

---

As one can see from the equation (1.1) solving this problem leads to a dynamically changed separating plane which depends on input vector $\mathbf{x}$ and on function set $\mathbf{f}(\cdot)$ used.

When solving the problem under the restrictions (1.2) we search special supporting hyperplanes to comply with the constraints which are transformed into the equalities (2.4), and we shall demonstrate that on a simple computational example. We shall show that for these planes and only for them Lagrange multipliers are not equal to zero. According to this we propose to name this method as Supporting Hyperplane Machine (SHM).

The functional (1.3) to be minimized is similar to the expressions used recently in SVM methods [2, 3, 4, 5]. Within these approaches two non-parallel planes are to be found out which lie near from the prescribed classes. One can see that the search of the method proposed by us is not confined with a pair of the planes.

When elaborating an efficient computational algorithm we have paid attention to the classifier ensembles used together with SVM [6]. Using popular methods [7, 8] we unfortunately have not yet created an algorithm with the speed to be higher than that of the well-known SMO [9, 10] for ordinary SVM e.g. on the base of LIBSVM [11].

So we present to the reader this manuscript as a detailed description of the SHM method version proposed by us.

## 2. Preliminary remarks.

Assume that there are a function set $\{f_t\colon \mathbb{R}^m \longrightarrow \mathbb{R}, \mathbf{x} \longmapsto \mathrm{y}\}_{t=1}^{Z}$ to perform in total the transformation $\mathbf{x} \longmapsto \mathbf{y}$, where $\mathbf{y} \in \mathbb{R}^Z$ and there is also a training set $\{(\mathbf{x}_i, d_i) | \mathbf{x}_i \in \mathbb{R}^m, d_i \in \{-1,1\}\}_{i=1}^{N}$, where $\mathbf{x}_i$ is the input image of the $i$-th example and $d_i$ is the corresponding desirable response to be independent of $\mathbf{x} \longmapsto \mathbf{y}$. When applying the transformation $\mathbf{x} \longmapsto \mathbf{y}$ to this set we shall get the training set $\{(\mathbf{x}_i, \mathbf{y}_i, d_i) | \mathbf{x}_i \in \mathbb{R}^m, \mathbf{y}_i \in \mathbb{R}^Z, d_i \in \{-1,1\}\}_{i=1}^{N}$. Let us set the problem to construct the optimal hyperplane family in $\mathbb{R}^Z$

$$\mathbf{x}^{\mathrm{T}}\mathbf{W}\mathbf{y} + \mathbf{w}_0^{\mathrm{T}}\mathbf{x} + b = 0, \tag{2.1}$$

which goes in the vicinity of the vector subsets $\mathbf{y}_i \in \mathbb{R}^Z$ for which it is true either $d_i = +1$ or $d_i = -1$ so that the constraints are valid

$$\begin{aligned} \mathbf{x}_i^{\mathrm{T}}\mathbf{W}\mathbf{y}_i + \mathbf{w}_0^{\mathrm{T}}\mathbf{x}_i + b \geq 1 \text{ for } d_i = +1 \\ \mathbf{x}_i^{\mathrm{T}}\mathbf{W}\mathbf{y}_i + \mathbf{w}_0^{\mathrm{T}}\mathbf{x}_i + b \leq -1 \text{ for } d_i = -1 \end{aligned} \tag{2.2}$$

It is obvious that the algebraic distance from some $\mathbf{y}$ point to a hyperplane belonging to the family (2.1) is

$$r = \frac{\mathbf{x}^{\mathrm{T}}\mathbf{W}\mathbf{y} + \mathbf{w}_0^{\mathrm{T}}\mathbf{x} + b}{\|[\mathbf{w}_0^{\mathrm{T}}\mathbf{x},\ \ \mathbf{x}^{\mathrm{T}}\mathbf{W}]\|} \tag{2.3}$$

Some points $\mathbf{y}_i$ from the training set $\{(\mathbf{x}_i, \mathbf{y}_i, d_i) | \mathbf{x}_i \in \mathbb{R}^m, \mathbf{y}_i \in \mathbb{R}^Z, d_i \in \{-1,1\}\}_{i=1}^{N}$ for which the constraints (2.2) are satisfied with the equality sign, lie on the corresponding supporting hyperplanes of subsets of the set $\{(\mathbf{x}_i, \mathbf{y}_i, d_i) | \mathbf{x}_i \in \mathbb{R}^m, \mathbf{y}_i \in \mathbb{R}^Z, d_i = -1\}$ and of subsets of the set $\{(\mathbf{x}_i, \mathbf{y}_i, d_i) | \mathbf{x}_i \in \mathbb{R}^m, \mathbf{y}_i \in \mathbb{R}^Z, d_i = +1\}$

$$\begin{aligned} \mathbf{x}_i^{\mathrm{T}}\mathbf{W}\mathbf{y}_i + \mathbf{w}_0^{\mathrm{T}}\mathbf{x}_i + b = 1 \text{ for } d_i = +1 \\ \mathbf{x}_i^{\mathrm{T}}\mathbf{W}\mathbf{y}_i + \mathbf{w}_0^{\mathrm{T}}\mathbf{x}_i + b = -1 \text{ for } d_i = -1 \end{aligned} \tag{2.4}$$

The algebraic distance from the supporting hyperplane of (2.4) to the hyperplane of (2.1) is equal to

$$r_i = \begin{cases} \dfrac{1}{\left\|[\mathbf{w}_0^{\mathrm{T}}\mathbf{x}_i,\ \ \mathbf{x}_i^{\mathrm{T}}\mathbf{W}]\right\|}, \text{if } d_i = +1 \\ -\dfrac{1}{\left\|[\mathbf{w}_0^{\mathrm{T}}\mathbf{x}_i,\ \ \mathbf{x}_i^{\mathrm{T}}\mathbf{W}]\right\|}, \text{if } d_i = -1 \end{cases} \tag{2.5}$$

It is obvious that $r_i$ distance maximization corresponds in each specific case to minimizing of Euclidean norm of the vector $[\mathbf{w}_0^{\mathrm{T}}\mathbf{x}_i,\ \ \mathbf{x}_i^{\mathrm{T}}\mathbf{W}]$. Then the problem to search the optimal hyperplane family (2.1) to provide the maximal distance of the hyperplanes (2.4) from the family (2.1) may be reduced to minimizing the functional $\sum_{i=1}^{N} \left\| [\mathbf{w}_0^{\mathrm{T}}\mathbf{x}_i, \quad \mathbf{x}_i^{\mathrm{T}}\mathbf{W}] \right\|^2$ under valid constraints (2.2). Let us denote this problem as the Linear SHM for certainty reasons.

## 3. The Linear SHM.

Solution of the Linear SHM problem is in whole similar to that of support vector machines. At first we establish the strict formulation of the problem.

---

So for the given training set $\{(\mathbf{x}_i, \mathbf{y}_i, d_i) | \mathbf{x}_i \in \mathbb{R}^m, \mathbf{y}_i \in \mathbb{R}^Z, d_i \in \{-1,1\}\}_{i=1}^{N}$ it is required to find out the weight factor matrix **W**, the weight factor vector $\mathbf{w}_0$ and the threshold $b$ which satisfy the constraints

$$d_i(\mathbf{x}_i^{\mathrm{T}}\mathbf{W}\mathbf{y}_i + \mathbf{w}_0^{\mathrm{T}}\mathbf{x}_i + b) \geq 1 \text{ for any } i = 1,2,\ldots,N \tag{3.1}$$

and minimize the functional

$$\Phi(\mathbf{W}, \mathbf{w}_0) = \frac{1}{2}\sum_{i=1}^{N} \left[ (\mathbf{x}_i^{\mathrm{T}}\mathbf{W})(\mathbf{x}_i^{\mathrm{T}}\mathbf{W})^{\mathrm{T}} + (\mathbf{w}_0^{\mathrm{T}}\mathbf{x}_i)^2 \right] \tag{3.2}$$

---

The formulated problem is essentially the task to search the constrained minimum of the function (3.2) under the constraints (3.1). Now we reformulate this initial problem as the task to search the saddle point $(\widehat{\mathbf{W}}, \widehat{\mathbf{w}}_0, \hat{b}, \widehat{\boldsymbol{\alpha}})$ of Lagrange function $L(\mathbf{W}, \mathbf{w}_0, b, \boldsymbol{\alpha})$ according to Karush - Kuhn - Tucker *optimality conditions* and *the saddle point theorem* [12].

Define at first the Lagrange function

$$L(\mathbf{W}, \mathbf{w}_0, b, \boldsymbol{\alpha}) = \frac{1}{2}\sum_{i=1}^{N} \left[ (\mathbf{x}_i^{\mathrm{T}}\mathbf{W})(\mathbf{x}_i^{\mathrm{T}}\mathbf{W})^{\mathrm{T}} + (\mathbf{w}_0^{\mathrm{T}}\mathbf{x}_i)^2 \right] - \sum_{i=1}^{N} \alpha_i \left[ d_i(\mathbf{x}_i^{\mathrm{T}}\mathbf{W}\mathbf{y}_i + \mathbf{w}_0^{\mathrm{T}}\mathbf{x}_i + b) - 1 \right] \tag{3.3}$$

Then calculate sequentially the partial derivatives $L(\mathbf{W}, \mathbf{w}_0, b, \boldsymbol{\alpha})$ with respect to **W**, $\mathbf{w}_0$ and $b$ according to the stationarity conditions and set them equal to zero.

$$\frac{\partial L(\mathbf{W}, \mathbf{w}_0, b, \boldsymbol{\alpha})}{\partial w_{pt}} = \sum_{i=1}^{N} \left( \sum_{l=1}^{m} w_{lt} x_{li} \right) x_{pi} - \sum_{i=1}^{N} \alpha_i d_i y_{ti} x_{pi} = 0, \tag{3.4}$$

for any $p = 1,2, \dots, m$ and $t = 1,2, \dots, Z$

$$\frac{\partial L(\mathbf{W}, \mathbf{w}_0, b, \boldsymbol{\alpha})}{\partial w_{p0}} = \sum_{i=1}^{N} \left( \sum_{l=1}^{m} w_{l0} x_{li} \right) x_{pi} - \sum_{i=1}^{N} \alpha_i d_i x_{pi} = 0, \qquad \text{for any } p = 1,2, \dots, m \tag{3.5}$$

$$\frac{\partial L(\mathbf{W}, \mathbf{w}_0, b, \boldsymbol{\alpha})}{\partial b} = \sum_{i=1}^{N} \alpha_i d_i = 0 \tag{3.6}$$

Regroup the multipliers in the expression (3.4).

$$\sum_{i=1}^{N} w_{lt} \left( \sum_{l=1}^{m} x_{pi} x_{li} \right) = \sum_{i=1}^{N} \alpha_i d_i y_{ti} x_{pi} \tag{3.7}$$

Rewriting (3.7) in the vector presentation one can obtain a family of systems of equations

$$(\mathbf{X}\mathbf{X}^{\mathrm{T}}) \mathbf{w}_t = \sum_{i=1}^{N} \alpha_i d_i y_{ti} \mathbf{x}_i \tag{3.8}$$

Here $\mathbf{w}_t$ denotes the corresponding column of the matrix $\mathbf{W}$. The necessity to solve the system (3.8) implies the restriction to the matrix $\mathbf{XX}^{\mathrm{T}}$. Its determinant is to be non-zero. Based on experience reasons one can expect the matrix $\mathbf{XX}^{\mathrm{T}}$ to be of full (numerical) rank. If this is not the case one should use one of the regularization methods [13].

Solutions for the family of the systems (3.8) are written either separately for each vector $\mathbf{w}_t$

$$\mathbf{w}_t = (\mathbf{X}\mathbf{X}^{\mathrm{T}})^{-} \sum_{i=1}^{N} \alpha_i d_i y_{ti} \mathbf{x}_i, \tag{3.9}$$

or in whole as the matrix equation

$$\mathbf{W} = (\mathbf{X}\mathbf{X}^{\mathrm{T}})^{-} \sum_{i=1}^{N} \alpha_i d_i \mathbf{x}_i \otimes \mathbf{y}_i^{\mathrm{T}}, \tag{3.10}$$

where the symbol $\otimes$ denotes Kronecker product and in this case defines the matrix product of the vector column $\mathbf{x}_i$ by the vector row $\mathbf{y}_i^{\mathrm{T}}$.

Similarly one can solve the expression (3.5) to obtain

$$\mathbf{w}_0 = (\mathbf{XX}^{\mathrm{T}})^{-}\sum_{i=1}^{N}\alpha_i d_i \mathbf{x}_i, \tag{3.11}$$

The expression (3.6) may be rewritten without the left-hand side

$$\sum_{i=1}^{N}\alpha_i d_i = 0 \tag{3.12}$$

One can express the complementary slackness condition

$$\alpha_i\left[d_i\left(\mathbf{x}_i^{\mathrm{T}}\mathbf{W}\mathbf{y}_i + \mathbf{w}_0^{\mathrm{T}}\mathbf{x}_i\ + b\right) - 1\right] = 0 \text{ for any } i = 1, 2, \dots, N \tag{3.13}$$

This relation means that only Lagrange multipliers for which the restrictions (3.1) are strictly fulfilled, have non-zero values.

The dual feasibility condition implies the following constraint to the Lagrange multipliers

$$\alpha_i \geq 0 \text{ for any } i = 1, 2, \dots, N \tag{3.14}$$

The original constraints (3.1) satisfy primal feasibility condition.

As the functional (3.2) is convex and the constraints (3.1) satisfy Slater condition, it is possible according to *the strong duality theorem* [12] to formulate the duality problem as follows

$$\min_{\mathbf{W},\mathbf{w_0}} L\left(\mathbf{W}, \mathbf{w_0}, \hat{b}, \widehat{\boldsymbol{\alpha}}\right) = \max_{\boldsymbol{\alpha}\geq\mathbf{0}} Q(\boldsymbol{\alpha}) \tag{3.15}$$

To calculate the function $Q(\boldsymbol{\alpha})$ one can substitute the relations (3.10-11) into (3.3) to obtain

$$Q(\boldsymbol{\alpha}) = \frac{1}{2}\sum_{i=1}^{N}\left[\left\|\mathbf{x}_i^{\mathrm{T}}\left((\mathbf{XX}^{\mathrm{T}})^{-}\sum_{j=1}^{N}\alpha_j d_j \mathbf{x}_j \otimes \mathbf{y}_j^{\mathrm{T}}\right)\right\|^2 + \left(\left((\mathbf{XX}^{\mathrm{T}})^{-}\sum_{j=1}^{N}\alpha_j d_j \mathbf{x}_j\right)^{\mathrm{T}}\mathbf{x}_i\right)^2\right] \tag{3.16}$$
$$-\sum_{i=1}^{N}\alpha_i\left[d_i\left(\mathbf{x}_i^{\mathrm{T}}\left((\mathbf{XX}^{\mathrm{T}})^{-}\sum_{j=1}^{N}\alpha_j d_j \mathbf{x}_j \otimes \mathbf{y}_j^{\mathrm{T}}\right)\mathbf{y}_i\right.\right.$$
$$\left.\left.+\left((\mathbf{XX}^{\mathrm{T}})^{-}\sum_{j=1}^{N}\alpha_j d_j \mathbf{x}_j\right)^{\mathrm{T}}\mathbf{x}_i\ + b\right) - 1\right]$$

The expression (3.16) after removal of all brackets and adductions similar terms may be rewritten as follows

$$Q(\boldsymbol{\alpha}) = \frac{1}{2}\sum_{i=1}^{N}\left[\sum_{t=1}^{Z}\left(\sum_{j=1}^{N}\alpha_j d_j y_{tj}\left((\mathbf{X}\mathbf{X}^{\mathrm{T}})^{-}\mathbf{x}_j\right)^{\mathrm{T}}\mathbf{x}_i\right)^2 + \left(\sum_{j=1}^{N}\alpha_j d_j\left((\mathbf{X}\mathbf{X}^{\mathrm{T}})^{-}\mathbf{x}_j\right)^{\mathrm{T}}\mathbf{x}_i\right)^2\right] \tag{3.17}$$
$$-\sum_{i=1}^{N}\alpha_i d_i\sum_{t=1}^{Z}y_{ti}\left(\sum_{j=1}^{N}\alpha_j d_j y_{tj}\left((\mathbf{X}\mathbf{X}^{\mathrm{T}})^{-}\mathbf{x}_j\right)^{\mathrm{T}}\mathbf{x}_i\right)$$
$$-\sum_{i=1}^{N}\alpha_i d_i\left(\sum_{j=1}^{N}\alpha_j d_j\left((\mathbf{X}\mathbf{X}^{\mathrm{T}})^{-}\mathbf{x}_j\right)^{\mathrm{T}}\mathbf{x}_i\right) - \sum_{i=1}^{N}\alpha_i d_i b + \sum_{i=1}^{N}\alpha_i$$

Introduce the following notation

$$g_{ji} = \left((\mathbf{X}\mathbf{X}^{\mathrm{T}})^{-}\mathbf{x}_j\right)^{\mathrm{T}}\mathbf{x}_i \tag{3.18}$$

Rewrite the function $Q(\boldsymbol{\alpha})$ taking into account (3.12) and (3.18)

$$Q(\boldsymbol{\alpha}) = \frac{1}{2}\sum_{i=1}^{N}\left[\sum_{t=1}^{Z}\left(\sum_{j=1}^{N}\alpha_j d_j y_{tj} g_{ji}\right)^2 + \left(\sum_{j=1}^{N}\alpha_j d_j g_{ji}\right)^2\right] \tag{3.19}$$
$$-\sum_{i=1}^{N}\alpha_i d_i\sum_{t=1}^{Z}y_{ti}\left(\sum_{j=1}^{N}\alpha_j d_j y_{tj} g_{ji}\right) - \sum_{i=1}^{N}\alpha_i d_i\left(\sum_{j=1}^{N}\alpha_j d_j g_{ji}\right) + \sum_{i=1}^{N}\alpha_i$$

It is necessary to note here that $g_{ji}$ are the elements of a matrix

$$\mathbf{G} = \mathbf{X}^{\mathrm{T}}(\mathbf{X}\mathbf{X}^{\mathrm{T}})^{-}\mathbf{X} \tag{3.20}$$

The matrix **G** is an orthogonal projector and has the idempotency and symmetry properties i.e. the following is valid

$$g_{jr} = \sum_{i=1}^{N} g_{ji} g_{ri} \text{ и } g_{ji} = g_{ij} \tag{3.21}$$

After removal of brackets in (3.19) we receive

$$Q(\boldsymbol{\alpha}) = \frac{1}{2}\sum_{i=1}^{N}\left[\sum_{t=1}^{Z}\sum_{j=1}^{N}\sum_{r=1}^{N}\alpha_j\alpha_r d_j d_r y_{tj} y_{tr} g_{ji} g_{ri} + \sum_{t=1}^{Z}\sum_{j=1}^{N}\sum_{r=1}^{N}\alpha_j\alpha_r d_j d_r g_{ji} g_{ri}\right] - \sum_{r=1}^{N}\sum_{t=1}^{Z}\sum_{j=1}^{N}\alpha_j\alpha_r d_j d_r y_{tj} y_{tr} g_{jr} - \sum_{r=1}^{N}\sum_{t=1}^{Z}\sum_{j=1}^{N}\alpha_j\alpha_r d_j d_r g_{jr} + \sum_{r=1}^{N}\alpha_r \quad (3.22)$$

Using (3.21) one can finally write the function Q as follows

$$Q(\boldsymbol{\alpha}) = -\frac{1}{2}\sum_{j=1}^{N}\sum_{r=1}^{N}\alpha_j\alpha_r d_j d_r\left(\mathbf{y}_j^{\mathrm{T}}\mathbf{y}_r + 1\right)g_{jr} + \sum_{r=1}^{N}\alpha_r \quad (3.23)$$

So we can now formulate the dual problem as the quadratic programming problem.

---

For the given training set $\{(\mathbf{x}_i, \mathbf{y}_i, d_i) | \mathbf{x}_i \in \mathbb{R}^m, \mathbf{y}_i \in \mathbb{R}^z, d_i \in \{-1,1\}\}_{i=1}^{N}$ one needs to find out the optimal Lagrange multipliers $\{\hat{\alpha}_i\}_{i=1}^{N}$ to maximize the target function

$$Q(\boldsymbol{\alpha}) = -\frac{1}{2}\sum_{i=1}^{N}\sum_{j=1}^{N}\alpha_i\alpha_j d_i d_j\left(\mathbf{y}_i^{\mathrm{T}}\mathbf{y}_j + 1\right)g_{ij} + \sum_{i=1}^{N}\alpha_i \quad (3.24)$$

under the following constraints

$$\hat{\alpha}_i \geq 0 \text{ for any } i = 1, 2, \dots, N \quad (3.25)$$
$$\sum_{i=1}^{N}\hat{\alpha}_i d_i = 0$$
$$\det(\mathbf{X}\mathbf{X}^{\mathrm{T}}) \neq 0$$

---

The optimal threshold $\hat{b}$ may be found out from the expression (2.1) e.g. as follows

$$\hat{b} = -\,\mathbf{x}^{\mathrm{T}}\widehat{\mathbf{W}}\mathbf{y} - \widehat{\mathbf{w}}_0^{\mathrm{T}}\mathbf{x}\,, \quad (3.26)$$

where $\{\mathbf{x}, \mathbf{y} | \mathrm{d} = +1\}$ is an example from the training set.

## 4. The binary classification problem.

As the optimal hyperplane family parameters are known now on the training set, we can create the corresponding binary classifications to recognize unknown examples $\mathbf{x}$ which use additional information provided by the function set $\{f_t: \mathbb{R}^m \longrightarrow \mathbb{R}, \mathbf{x} \longmapsto \mathrm{y}\}_{t=1}^{z}$. For this purpose one can substitute the expressions (3.10-11) into the left-hand side of the equation (2.1) to get the following function

$$h(\mathbf{x},\mathbf{y}) = \mathbf{x}^{\mathrm{T}}\left((\mathbf{X}\mathbf{X}^{\mathrm{T}})^{-}\sum_{i=1}^{N}\hat{\alpha}_i d_i \mathbf{x}_i \otimes \mathbf{y}_i^{\mathrm{T}}\right)\mathbf{y} + \left((\mathbf{X}\mathbf{X}^{\mathrm{T}})^{-}\sum_{i=1}^{N}\hat{\alpha}_i d_i \mathbf{x}_i\right)^{\mathrm{T}}\mathbf{x} + \hat{b} \tag{4.1}$$

$$= (\mathbf{X}\mathbf{X}^{\mathrm{T}})^{-}\left(\sum_{i=1}^{N}\hat{\alpha}_i d_i (\mathbf{y}_i^{\mathrm{T}}\mathbf{y} + 1)\mathbf{x}_i\right)^{\mathrm{T}} + \hat{b}$$

## 5. Compressed presentation of Linear SHM.

Let the training set $\{(\mathbf{x}_i, \mathbf{y}_i, d_i) | \mathbf{x}_i \in \mathbb{R}^m, \mathbf{y}_i \in \mathbb{R}^z, d_i \in \{-1,1\}\}_{i=1}^{N}$ exist in which each $\mathbf{y}_i$ is a result of transformation by means of the function set $\{f_t: \mathbb{R}^m \longrightarrow \mathbb{R}, \mathbf{x} \longmapsto \mathrm{y}\}_{t=1}^{z}$ on the input vector $\mathbf{x}_i$ and $d_i \in \{-1,1\}$ presents the desired response independent of $\mathbf{x} \longmapsto \mathbf{y}$. Set the task to construct the optimal hyperplane family in $\mathbb{R}^z$

$$\mathbf{x}^{\mathrm{T}}\mathbf{W}\mathbf{y} + \mathbf{w}_0^{\mathrm{T}}\mathbf{x} + b = 0 \tag{5.1}$$

as follows

$$\min_{\mathbf{W},\mathbf{w}_0} \Phi(\mathbf{W},\mathbf{w}_0) = \frac{1}{2}\sum_{i=1}^{N}\left[(\mathbf{x}_i^{\mathrm{T}}\mathbf{W})(\mathbf{x}_i^{\mathrm{T}}\mathbf{W})^{\mathrm{T}} + (\mathbf{w}_0^{\mathrm{T}}\mathbf{x}_i)^2\right] \tag{5.2}$$

$$\text{subject to } \forall i\ d_i(\mathbf{x}_i^{\mathrm{T}}\mathbf{W}\mathbf{y}_i + \mathbf{w}_0^{\mathrm{T}}\mathbf{x}_i + b) \geq 1$$

Due to Karush - Kuhn - Tucker *optimality conditions* and *the saddle point theorem* we reformulate (5.2) as the problem to search the saddle point of Lagrange function.

$$L(\widehat{\mathbf{W}}, \widehat{\mathbf{w}_0}, \hat{b}, \widehat{\boldsymbol{\alpha}}) = \min_{\mathbf{W},\mathbf{w}_0}\max_{\boldsymbol{\alpha}\geq\mathbf{0}}\left\{\frac{1}{2}\sum_{i=1}^{N}\left[(\mathbf{x}_i^{\mathrm{T}}\mathbf{W})(\mathbf{x}_i^{\mathrm{T}}\mathbf{W})^{\mathrm{T}} + (\mathbf{w}_0^{\mathrm{T}}\mathbf{x}_i)^2\right] \right. \tag{5.3}$$

$$\left. - \sum_{i=1}^{N}\alpha_i\left[d_i(\mathbf{x}_i^{\mathrm{T}}\mathbf{W}\mathbf{y}_i + \mathbf{w}_0^{\mathrm{T}}\mathbf{x}_i + b) - 1\right]\right\}$$

According to Karush - Kuhn - Tucker stationarity condition one can derive the following equations

$$\mathbf{W} = (\mathbf{X}\mathbf{X}^{\mathrm{T}})^{-}\sum_{i=1}^{N}\alpha_i d_i \mathbf{x}_i \otimes \mathbf{y}_i^{\mathrm{T}} \tag{5.4}$$

$$\mathbf{w}_0 = (\mathbf{X}\mathbf{X}^{\mathrm{T}})^{-}\sum_{i=1}^{N}\alpha_i d_i \mathbf{x}_i$$

$$\sum_{i=1}^{N}\alpha_i d_i = 0$$

Using the equations (5.3-4) and based on *the strong duality theorem* one can reformulate the problem (5.2) as the dual quadratic programming problem which covers only dual variables.

$$Q(\widehat{\boldsymbol{\alpha}}) = \max_{\alpha_i}\left\{-\frac{1}{2}\sum_{i=1}^{N}\sum_{j=1}^{N}\alpha_i\alpha_j d_i d_j\left(\mathbf{y}_i^{\mathrm{T}}\mathbf{y}_j + 1\right)g_{ij} + \sum_{i=1}^{N}\alpha_i\right\} \tag{5.5}$$

$$\text{subject to}\quad\begin{cases}\forall i\ \alpha_i \geq 0\\ \sum_{i=1}^{N}\alpha_i d_i = 0\\ \mathbf{G} = \mathbf{X}^{\mathrm{T}}(\mathbf{X}\mathbf{X}^{\mathrm{T}})^{-}\mathbf{X}\\ \det(\mathbf{X}\mathbf{X}^{\mathrm{T}}) \neq 0\end{cases}$$

## 6. A calculation example.

To demonstrate the method we shall consider on the plane a simple binary classification problem with the training set of 16 points using two functions specified in the table form. Besides the figures and table in the text all the initial and intermediate values together with the final results which are necessary for calculations are specified in Appendix A to make it easier to check them at any moment.

Let us consider the numbered set $\left\{(\mathbf{x}_i, d_i)\middle|\mathbf{x}_i \in \mathbb{R}^{m=2}, d_i \in \{-1,1\}\right\}_{i=1}^{N=16}$, shown in Fig. 1. Here the blue color denotes the «negative» examples and the red color denotes the «positive» ones which for convenience reasons correspond to odd and even numbers.

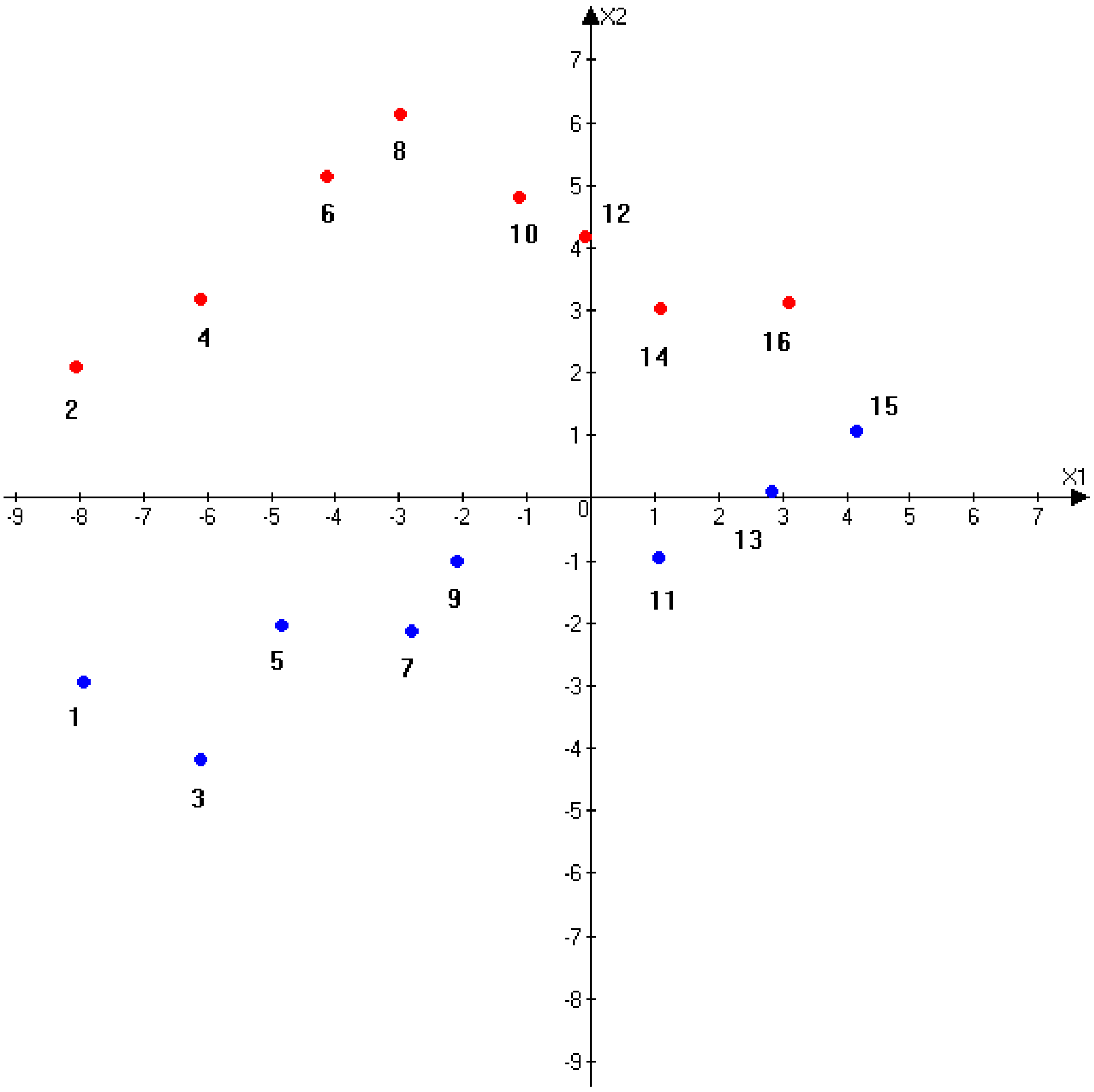

Fig. 1. Numbered training set of two classes.

We apply two tabular functions $f_1(\cdot)$ and $f_2(\cdot)$ to this set to perform together the transformation $\mathbf{x} \mapsto \mathbf{y}$. The result of the transformation as the numbered set $\{(\mathbf{y}_i, d_i) | \mathbf{y}_i \in \mathbb{R}^{m=2}, d_i \in \{-1,1\}\}_{i=1}^{N=16}$ may be seen in Fig. 2. One can see from this Figure that both functions perform separation of the learning examples to «negative» and «positive» ones with only partial success. So the function $f_1(\cdot)$ corresponding to the output values $y_{1i}$ successfully performs separation of the examples from number 1 till 8 inclusive, and the function $f_2(\cdot)$ corresponding to the output values $y_{2i}$ provides separation from number 9 till 16.

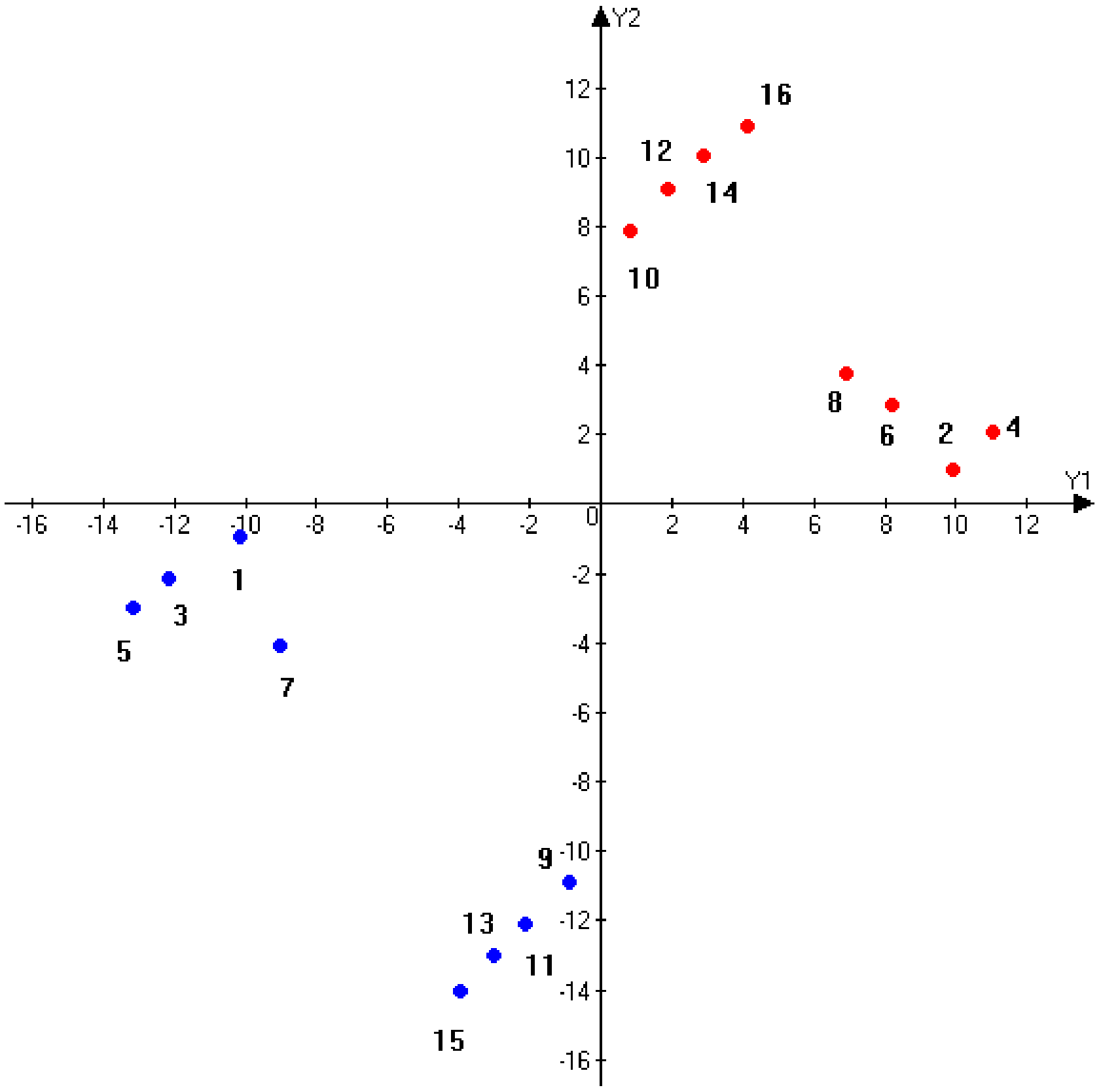

Fig. 2. The result of the transformation $\mathbf{x} \longmapsto \mathbf{y}$.

Now there is available the training set $\left\{(\mathbf{x}_i, \mathbf{y}_i, d_i) \middle| \mathbf{x}_i \in \mathbb{R}^{m=2}, \mathbf{y}_i \in \mathbb{R}^{Z=2}, d_i \in \{-1,1\}\right\}_{i=1}^{N=16}$. Based on this set and by means of our SHM approach we shall get a new recognizing function $\mathrm{h}(\mathbf{x}, \mathbf{y}) = \mathbf{x}^{\mathrm{T}}\mathbf{W}\mathbf{y} + \mathbf{w}_0^{\mathrm{T}}\mathbf{x} + b$

Construct the following algorithm.

**Algorithm 1**. The SHM algorithm.

---

1: Calculate the covariance matrix $\mathbf{X}\mathbf{X}^{\mathrm{T}}$.

2: Calculate the matrix $(\mathbf{X}\mathbf{X}^{\mathrm{T}})^{-}$. (Use regularization if needed).

3: Compute the orthogonal projector $\mathbf{G} = \mathbf{X}^{\mathrm{T}}(\mathbf{X}\mathbf{X}^{\mathrm{T}})^{-}\mathbf{X}$.

4: Calculate the kernel trick matrix $\mathbf{K} = (\mathbf{Y}^{\mathrm{T}}\mathbf{Y} + 1)$.

5: Calculate the Hessian $\mathbf{H} = \mathbf{d}\mathbf{e}^{\mathrm{T}} \circ \mathbf{K} \circ \mathbf{G} \circ \mathbf{e}\mathbf{d}^{\mathrm{T}}$. (Here $\mathbf{e}$ denotes the unit vector and the notation $\circ$ stands for Hadamard product).

6: Solve the quadratic programming problem $\min_{\boldsymbol{\alpha}}\left\{\frac{1}{2}\boldsymbol{\alpha}^{\mathrm{T}}\mathbf{H}\boldsymbol{\alpha} - \mathbf{e}^{\mathrm{T}}\boldsymbol{\alpha}\right\}$ under the constraints $\forall i\ \alpha_i \geq 0$ and $\boldsymbol{\alpha}^{\mathrm{T}}\mathbf{d} = \mathbf{0}$.

7: Calculate the weight matrix $\widehat{\mathbf{W}} = (\mathbf{X}\mathbf{X}^{\mathrm{T}})^{-} \sum_{i=1}^{N} \hat{\alpha}_i d_i \mathbf{x}_i \otimes \mathbf{y}_i^{\mathrm{T}}$, the weight vector $\widehat{\mathbf{w}}_0 = (\mathbf{X}\mathbf{X}^{\mathrm{T}})^{-} \sum_{i=1}^{N} \hat{\alpha}_i d_i \mathbf{x}_i$ and the optimal threshold $\hat{b} = -\,\mathbf{x}^{\mathrm{T}}\widehat{\mathbf{W}}\mathbf{y} - \widehat{\mathbf{w}}_0^{\mathrm{T}}\mathbf{x}$, where $\{\mathbf{x}, \mathbf{y}, \mathrm{d} = +1\}$ is an example from the training set.

8: Obtain the recognizing function $\mathrm{h}(\mathbf{x}, \mathbf{y}) = \mathbf{x}^{\mathrm{T}}\widehat{\mathbf{W}}\mathbf{y} + \widehat{\mathbf{w}}_0^{\mathrm{T}}\mathbf{x} + \hat{b}$

---

After getting the recognizing function $\mathrm{h}(\mathbf{x}, \mathbf{y})$ under this algorithm and applying it to the training set we shall obtain the result summarized in Table 1.

Table 1. The result of applying of the recognizing function $\mathrm{h}(\mathbf{x}, \mathbf{y})$ to the training set (the values are rounded up).

| $i$ | 1 | 2 | 3 | 4 | 5 | 6 | 7 | 8 | 9 | 10 | 11 | 12 | 13 | 14 | 15 | 16 |
|---|---|---|---|---|---|---|---|---|---|---|---|---|---|---|---|---|
| $\mathrm{h}(\mathbf{x}_i, \mathbf{y}_i)$ | -2,5 | 1,0 | -2,7 | 2,7 | -1,0 | 5,2 | -1,9 | 5,7 | -1,0 | 1,0 | -3,0 | 1,2 | -2,0 | 1,9 | -1,0 | 3,9 |

One can see that the result may be similar to the result of an applying the classic SVM algorithm on the set $\left\{(\mathbf{y}_i, d_i)\middle|\mathbf{y}_i \in \mathbb{R}^{m=2}, d_i \in \{-1,1\}\right\}_{i=1}^{N=16}$. However we (it is worth to stress this once more!) use the training set $\left\{(\mathbf{x}_i, \mathbf{y}_i, d_i)\middle|\mathbf{x}_i \in \mathbb{R}^{m=2}, \mathbf{y}_i \in \mathbb{R}^{Z=2}, d_i \in \{-1,1\}\right\}_{i=1}^{N=16}$ in the proposed SHM method to involve the additional information contained in the transformation $\mathbf{x} \longmapsto \mathbf{y}$.

While performing the algorithm we obtain only some Lagrange multipliers to be nonzero (5 from 16 in our example – see Appendix A). Returning to Lagrange function (3.3) we notice that these nonzero multipliers $\alpha_i$ correspond to only the restrictions (2.2), which under the complementary slackness condition (3.13) transform into the equalities (2.4). These equalities correspond to the supporting hyperplanes. In the example considered these are 5 non-parallel segments of the straight lines shown in Fig. 3.

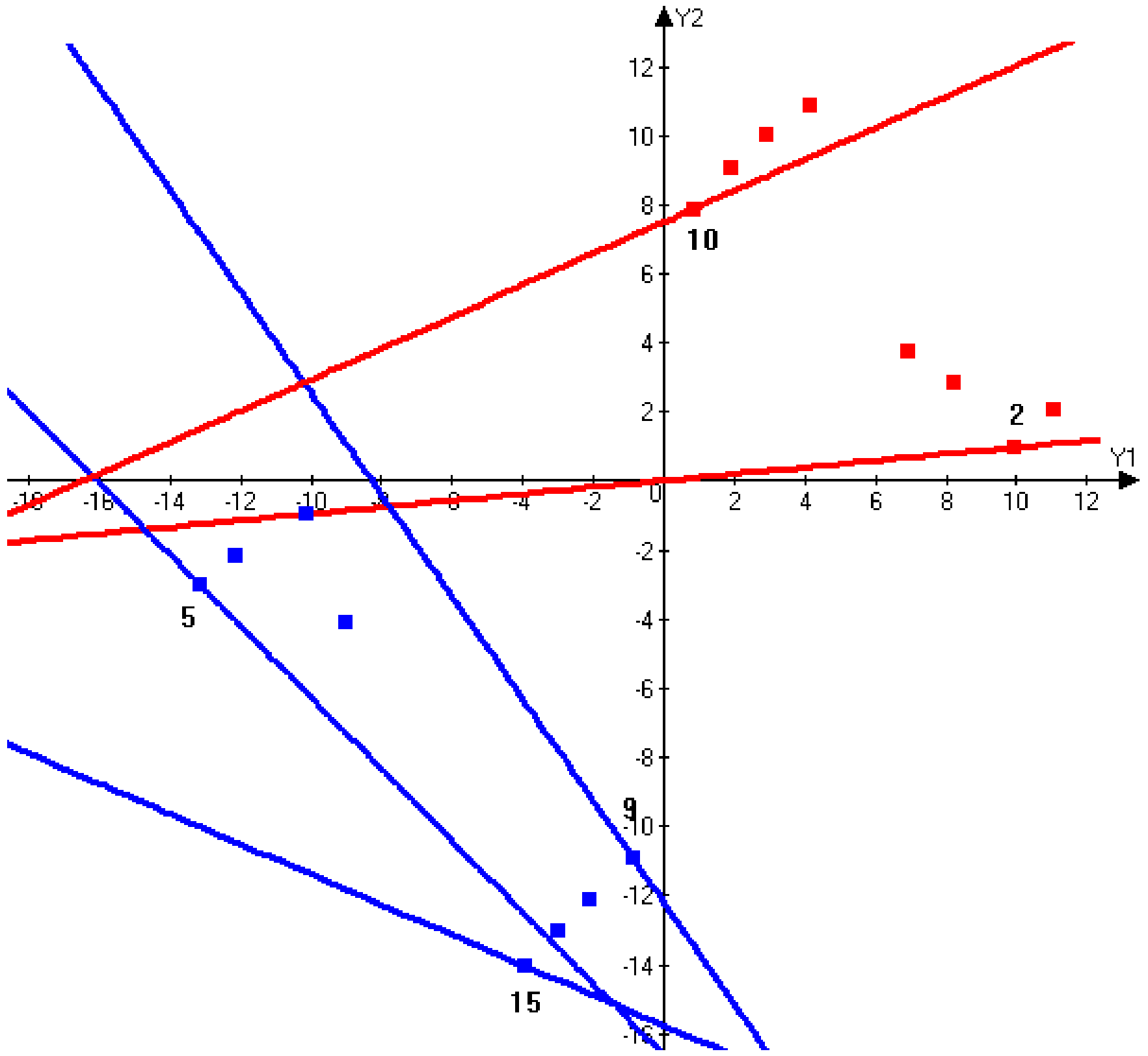

Fig. 3. The supporting hyperplanes for the training set considered.

Remind which are the straight lines on the plane to correspond to the supporting hyperplanes from (2.4) in $\mathbb{R}^2$. It is well-known from the elementary geometry that the general equation of the straight line in the plane is $Ay_1 + By_2 + C = 0$. In accordance with the equation the factors $A, B$ and $C$ are given for a specific line as follows:

$$[A, \quad B] = \mathbf{x}_i^{\mathrm{T}} \widehat{\mathbf{W}} \qquad (6.1)$$
$$C = \widehat{\mathbf{w}}_0^{\mathrm{T}} \mathbf{x}_i + \hat{b} - d_i$$

To understand better the example and the algorithm one can use the prepared script in Octave language [14] from Appendix B directly in GNU Octave environment, after translating into MATLAB or any other programming language. The script allows to obtain all the intermediate and final results which are presented in this passage and in Appendix A for the training set, including the results which are essential to construct the supporting hyperplanes. Upon checking the script for the presented example and comparing it with our results in Appendix A, one can easily use it with for own data.

## 7. Nonlinear SHM.

Consider the situation when besides the function set $\{f_t : \mathbb{R}^m \longrightarrow \mathbb{R}, \mathbf{x} \longmapsto \mathrm{y}\}_{t=1}^{z}$, the initial set of the learning examples $\{(\mathbf{x}_i, d_i) | \mathbf{x}_i \in \mathbb{R}^m, d_i \in \{-1,1\}\}_{i=1}^{N}$ there exists also a set of nonlinear transformations $\{\varphi_k(\mathbf{y})\}_{k=1}^{E}$ for which only the kernel trick is known. Using the kernel trick $K(\mathbf{y}, \mathbf{y}')$ we reformulate the problem (5.2) to search the optimal hyperplane family (5.1) as follows.

Let the training set $\{(\mathbf{x}_i, \boldsymbol{\varphi}(\mathbf{y}_i), d_i) | \mathbf{x}_i \in \mathbb{R}^m, \mathbf{y}_i \in \mathbb{R}^z, \boldsymbol{\varphi}(\mathbf{y}_i) \in \mathbb{R}^E, d_i \in \{-1,1\}\}_{i=1}^{N}$ exist to be obtained in the result of the transformation $\mathbf{x} \longmapsto \mathbf{y} \longmapsto \boldsymbol{\varphi}(\mathbf{y})$ for the initial set of examples $\{(\mathbf{x}_i, d_i) | \mathbf{x}_i \in \mathbb{R}^m, d_i \in \{-1,1\}\}_{i=1}^{N}$. Set the task to search the optimal hyperplane family in $\mathbb{R}^E$

$$\mathbf{x}^{\mathrm{T}}\mathbf{W}\boldsymbol{\varphi}(\mathbf{y}) + \mathbf{w}_0^{\mathrm{T}}\mathbf{x} + b = 0 \tag{7.1}$$

as follows

$$\min_{\mathbf{W},\mathbf{w}_0} \Phi(\mathbf{W}, \mathbf{w}_0) = \frac{1}{2}\sum_{i=1}^{N} \left[ (\mathbf{x}_i^{\mathrm{T}}\mathbf{W})(\mathbf{x}_i^{\mathrm{T}}\mathbf{W})^{\mathrm{T}} + (\mathbf{w}_0^{\mathrm{T}}\mathbf{x}_i)^2 \right] \tag{7.2}$$
$$\text{subject to } \forall i \; d_i(\mathbf{x}_i^{\mathrm{T}}\mathbf{W}\boldsymbol{\varphi}(\mathbf{y}_i) + \mathbf{w}_0^{\mathrm{T}}\mathbf{x}_i + b) \geq 1$$

We shall search the saddle point of Lagrange function as

$$L(\widehat{\mathbf{W}}, \widehat{\mathbf{w}_0}, \hat{b}, \widehat{\boldsymbol{\alpha}}) = \min_{\mathbf{W},\mathbf{w}_0} \max_{\boldsymbol{\alpha}\geq\mathbf{0}} \left\{ \frac{1}{2}\sum_{i=1}^{N} \left[ (\mathbf{x}_i^{\mathrm{T}}\mathbf{W})(\mathbf{x}_i^{\mathrm{T}}\mathbf{W})^{\mathrm{T}} + (\mathbf{w}_0^{\mathrm{T}}\mathbf{x}_i)^2 \right] - \sum_{i=1}^{N} \alpha_i \left[ d_i(\mathbf{x}_i^{\mathrm{T}}\mathbf{W}\boldsymbol{\varphi}(\mathbf{y}_i) + \mathbf{w}_0^{\mathrm{T}}\mathbf{x}_i + b) - 1 \right] \right\} \tag{7.3}$$

Write in the stationarity conditions

$$\mathbf{W} = (\mathbf{X}\mathbf{X}^{\mathrm{T}})^{-} \sum_{i=1}^{N} \alpha_i d_i \mathbf{x}_i \otimes (\boldsymbol{\varphi}(\mathbf{y}_i))^{\mathrm{T}} \tag{7.4}$$
$$\mathbf{w}_0 = (\mathbf{X}\mathbf{X}^{\mathrm{T}})^{-} \sum_{i=1}^{N} \alpha_i d_i \mathbf{x}_i$$
$$\sum_{i=1}^{N} \alpha_i d_i = 0$$

AND formulate the dual problem

$$Q(\hat{\boldsymbol{\alpha}}) = \max_{\alpha_i}\left\{-\frac{1}{2}\sum_{i=1}^{N}\sum_{j=1}^{N}\alpha_i\alpha_j d_i d_j\big(K(\mathbf{y}_i,\mathbf{y}_j)+1\big)g_{ij} + \sum_{i=1}^{N}\alpha_i\right\} \quad (7.5)$$

$$\text{subject to} \begin{cases} \forall i\ \alpha_i \geq 0 \\ \sum_{i=1}^{N}\alpha_i d_i = 0 \\ \mathbf{G} = \mathbf{X}^{\mathrm{T}}(\mathbf{X}\mathbf{X}^{\mathrm{T}})^{-}\mathbf{X} \\ \det(\mathbf{X}\mathbf{X}^{\mathrm{T}}) \neq 0 \end{cases}$$

The recognizing function will change respectively

$$h(\mathbf{x},\mathbf{y}) = (\mathbf{X}\mathbf{X}^{\mathrm{T}})^{-}\left(\sum_{i=1}^{N}\hat{\alpha}_i d_i(K(\mathbf{y}_i,\mathbf{y})+1)\mathbf{x}_i\right)^{\mathrm{T}} + \hat{b} \quad (7.6)$$

## 8. Soft margin case.

So far we have assumed that the hyperplane families of (5.1) and (7.1) satisfy the strict constraints of the problems with the forms (5.2) or (7.2) under a some optimal set of the parameters. Similar to the support vector machine it may be permitted within the method proposed by us to follow these constraints with admissible errors to minimize better the structural risk. For this purpose we reformulate the problem (7.2) as follows

$$\min_{\mathbf{W},\mathbf{w}_0}\Phi(\mathbf{W},\mathbf{w}_0,\boldsymbol{\xi}) = \frac{1}{2}\sum_{i=1}^{N}\left[\left(\mathbf{x}_i^{\mathrm{T}}\mathbf{W}\right)\left(\mathbf{x}_i^{\mathrm{T}}\mathbf{W}\right)^{\mathrm{T}} + (\mathbf{w}_0^{\mathrm{T}}\mathbf{x}_i)^2\right] + C\sum_{i=1}^{N}\xi_i \quad (8.1)$$

$$\text{subject to } \forall i \begin{cases} d_i(\mathbf{x}_i^{T}\mathbf{W}\boldsymbol{\varphi}(\mathbf{y}_i) + \mathbf{w}_0^{\mathrm{T}}\mathbf{x}_i + b) \geq 1-\xi_i \\ \xi_i \geq 0 \end{cases}$$

Then the problem to search the saddle point of Lagrange function takes the form

$$L\big(\widehat{\mathbf{W}},\widehat{\mathbf{w}_0},\hat{b},\hat{\boldsymbol{\alpha}}\big) = \min_{\mathbf{W},\mathbf{w}_0}\max_{\boldsymbol{\alpha}\geq\mathbf{0},\boldsymbol{\beta}\geq\mathbf{0}}\left\{\frac{1}{2}\sum_{i=1}^{N}\left[\left(\mathbf{x}_i^{\mathrm{T}}\mathbf{W}\right)\left(\mathbf{x}_i^{\mathrm{T}}\mathbf{W}\right)^{\mathrm{T}} + (\mathbf{w}_0^{\mathrm{T}}\mathbf{x}_i)^2\right] + C\sum_{i=1}^{N}\xi_i \right.$$
$$\left. - \sum_{i=1}^{N}\alpha_i\left[d_i\left(\mathbf{x}_i^{\mathrm{T}}\mathbf{W}\boldsymbol{\varphi}(\mathbf{y}_i) + \mathbf{w}_0^{\mathrm{T}}\mathbf{x}_i + b\right) - 1 + \xi_i\right] - \sum_{i=1}^{N}\beta_i\xi_i\right\} \quad (8.2)$$

Besides to the existing stationarity conditions and the complementary slackness conditions the following relations will be added

$$\forall i\ \alpha_i + \beta_i = C \qquad (8.3)$$
$$\forall i\ \beta_i \xi_i = 0$$

Now we can formulate the dual quadratic programming problem without the variables $\xi_i$

$$Q(\widehat{\boldsymbol{\alpha}}) = \max_{\alpha_i} \left\{ -\frac{1}{2} \sum_{i=1}^{N} \sum_{j=1}^{N} \alpha_i \alpha_j d_i d_j \left( K(\mathbf{y}_i, \mathbf{y}_j) + 1 \right) g_{ij} + \sum_{i=1}^{N} \alpha_i \right\} \qquad (8.4)$$

$$\text{subject to} \quad \begin{cases} \forall i\ 0 \le \alpha_i \le C \\ \sum_{i=1}^{N} \alpha_i d_i = 0 \\ \mathbf{G} = \mathbf{X}^{\mathrm{T}} (\mathbf{X}\mathbf{X}^{\mathrm{T}})^{-} \mathbf{X} \\ \det(\mathbf{X}\mathbf{X}^{\mathrm{T}}) \neq 0 \end{cases}$$

## 9. SHM and condensed SVD theorem.

According to the condensed SVD theorem [13] we shall present the decomposition of the orthogonal projector

$$\mathbf{G} = \mathbf{U}_r \mathbf{S}_r \mathbf{V}_r^{\mathrm{T}} \qquad (9.1)$$
$$\mathbf{U}_r = \mathbf{V}_r^{\mathrm{T}}$$
$$\forall q\ s_q = 1$$
$$r = rank(\mathbf{G})$$

The round-up errors and using of the regularization may actually lead to the situation when the equalities (9.1) may not be fulfilled exactly up to a certain digit. Then the matrix $\mathbf{U}_r \mathbf{S}_r \mathbf{V}_r^{\mathrm{T}}$ will be somewhat only the low-rank approximation for the matrix $\mathbf{G}$. Here we shall require strict fulfillment of the equalities (9.1).

It is obvious from (9.1) that

$$\forall i,j\ g_{i,j} = \sum_{q}^{rank(\mathbf{G})} u_{iq} u_{jq} \qquad (9.2)$$

Under (9.2) replace the variables

$$\forall i \; c_i = d_i \sum_{q}^{rank(\mathbf{G})} u_{iq} \tag{9.3}$$

Using this we write the problem (7.5) as follows

$$Q(\widehat{\boldsymbol{\alpha}}) = \max_{\alpha_i} \left\{ -\frac{1}{2} \sum_{i=1}^{N} \sum_{j=1}^{N} \alpha_i \alpha_j c_i c_j \left( K(\mathbf{y}_i, \mathbf{y}_j) + 1 \right) + \sum_{i=1}^{N} \alpha_i \right\} \tag{9.4}$$

$$\text{subject to} \quad \begin{cases} \forall i \; 0 \leq \alpha_i \leq C \\ \forall i \; c_i = d_i \sum_{q}^{r} u_{iq} \\ \sum_{i=1}^{N} \alpha_i d_i = 0 \\ r = rank(\mathbf{G}) \\ \mathbf{G} = \mathbf{X}^{\mathrm{T}} (\mathbf{X}\mathbf{X}^{\mathrm{T}})^{-} \mathbf{X} \\ \mathbf{G} = \mathbf{U}_r \mathbf{S}_r \mathbf{V}_r^{\mathrm{T}} \\ \det(\mathbf{X}\mathbf{X}^{\mathrm{T}}) \neq 0 \end{cases}$$

Appendix A. Initial, intermediate and final data for the computational example.

**X**

| | | | | | | | | | | | | | | | |
|---|---|---|---|---|---|---|---|---|---|---|---|---|---|---|---|
| -7,94 | -8,05 | -6,12 | -6,10 | -4,85 | -4,13 | -2,81 | -2,97 | -2,09 | -1,11 | 1,05 | -0,10 | 2,84 | 1,10 | 4,15 | 3,11 |
| -2,94 | 2,09 | -4,17 | 3,18 | -2,04 | 5,13 | -2,14 | 6,13 | -1,00 | 4,80 | -0,94 | 4,18 | 0,11 | 3,04 | 1,07 | 3,11 |

$\mathbf{d}^T$

| | | | | | | | | | | | | | | | |
|---|---|---|---|---|---|---|---|---|---|---|---|---|---|---|---|
| -1 | 1 | -1 | 1 | -1 | 1 | -1 | 1 | -1 | 1 | -1 | 1 | -1 | 1 | -1 | 1 |

**Y**

| | | | | | | | | | | | | | | | |
|---|---|---|---|---|---|---|---|---|---|---|---|---|---|---|---|
| -10,17 | 9,93 | -12,18 | 11,02 | -13,18 | 8,20 | -9,04 | 6,88 | -0,89 | 0,83 | -2,11 | 1,87 | -3,03 | 2,88 | -3,93 | 4,11 |
| -0,92 | 0,97 | -2,13 | 2,08 | -2,96 | 2,86 | -4,05 | 3,80 | -10,88 | 7,87 | -12,07 | 9,10 | -12,99 | 10,04 | -14,04 | 10,88 |

| $\boldsymbol{\alpha}$ | $\mathbf{X}^T\widehat{\mathbf{W}}$ | | $\mathbf{X}^T\widehat{\mathbf{w}_0} + \mathbf{e}b - \mathbf{d}$ | $h(\mathbf{x_i}, \mathbf{y_i})$ |
|---|---|---|---|---|
| 0,0000 | | | | -2,5363 |
| 2,2771 | 0,0792 | -0,8244 | 0,0132 | 1,0000 |
| 0,0000 | | | | -2,7258 |
| 0,0000 | | | | 2,6956 |
| 9,8557 | -0,1445 | -0,1395 | -2,3174 | -1,0000 |
| 0,0000 | | | | 5,1564 |
| 0,0000 | | | | -1,8939 |
| 0,0000 | | | | 5,7426 |
| 162,5579 | -0,0693 | -0,0470 | -0,5735 | -1,0000 |
| 20,1065 | 0,2738 | -0,6021 | 4,5112 | 1,0000 |
| 0,0000 | | | | -2,9927 |
| 0,0000 | | | | 1,2272 |
| 0,0000 | | | | -2,0000 |
| 0,0000 | | | | 1,8642 |
| 69,5705 | 0,0843 | 0,1925 | 3,0343 | -1,0000 |
| 0,0000 | | | | 3,9359 |

| $(\mathbf{XX}^T)^-$ | |
|---|---|
| 0,0033 | 0,0000 |
| 0,0000 | 0,0057 |

| **W** | |
|---|---|
| 0,0053 | 0,0743 |
| 0,0583 | -0,1083 |

| $\mathbf{w}_0$ |
|---|
| 0,1832 |
| 1,1905 |

| $b$ |
|---|
| 0,0000 |

$\mathbf{H} = \mathbf{d}\mathbf{e}^{\mathrm{T}} \circ \mathbf{K} \circ \mathbf{G} \circ \mathbf{e}\mathbf{d}^{\mathrm{T}}$

| | | | | | | | | | | | | | | | |
|---|---|---|---|---|---|---|---|---|---|---|---|---|---|---|---|
| 26,8997 | 17,7437 | 28,9396 | 12,1009 | 22,0536 | 2,0137 | 10,4844 | -1,6670 | 1,4237 | -0,7304 | -0,4026 | -1,7404 | -3,3240 | -2,9488 | -6,7840 | -6,7171 |
| 17,7437 | 24,1742 | 13,6963 | 22,6849 | 13,8653 | 14,7439 | 4,5218 | 11,2675 | 0,8005 | 1,4920 | -1,2517 | 1,5295 | -3,1035 | 0,3099 | -5,0428 | -2,3516 |
| 28,9396 | 13,6963 | 33,9459 | 6,5486 | 24,3550 | -3,9578 | 12,7606 | -7,6747 | 2,2890 | -2,3410 | 0,0541 | -3,9580 | -3,9018 | -5,1825 | -8,5312 | -9,7781 |
| 12,1009 | 22,6849 | 6,5486 | 23,0966 | 9,1645 | 17,3264 | 1,8950 | 14,6468 | 0,7557 | 2,9355 | -1,8268 | 3,1951 | -3,3035 | 1,7979 | -4,6161 | -0,4115 |
| 22,0536 | 13,8653 | 24,3550 | 9,1645 | 18,4519 | 0,8423 | 9,1436 | -2,2815 | 2,0094 | -1,2322 | -0,3878 | -2,3317 | -3,6927 | -3,4815 | -7,3924 | -7,2608 |
| 2,0137 | 14,7439 | -3,9578 | 17,3264 | 0,8423 | 15,8988 | -2,0276 | 15,1149 | -0,0154 | 4,7433 | -2,1441 | 5,2630 | -2,2048 | 3,9517 | -1,8626 | 3,1781 |
| 10,4844 | 4,5218 | 12,7606 | 1,8950 | 9,1436 | -2,0276 | 5,1200 | -3,5738 | 1,6612 | -1,8311 | 0,1166 | -2,6126 | -2,2222 | -3,0742 | -4,7684 | -5,3036 |
| -1,6670 | 11,2675 | -7,6747 | 14,6468 | -2,2815 | 15,1149 | -3,5738 | 15,3610 | -0,6495 | 6,5676 | -2,5840 | 7,1439 | -1,7138 | 5,6271 | -0,3435 | 5,4821 |
| 1,4237 | 0,8005 | 2,2890 | 0,7557 | 2,0094 | -0,0154 | 1,6612 | -0,6495 | 2,3971 | -1,6464 | -0,2599 | -2,2669 | -2,9159 | -2,7243 | -5,4216 | -4,6929 |
| -0,7304 | 1,4920 | -2,3410 | 2,9355 | -1,2322 | 4,7433 | -1,8311 | 6,5676 | -1,6464 | 8,6364 | -2,8538 | 8,5182 | -0,8288 | 6,5086 | 1,4911 | 6,5830 |
| -0,4026 | -1,2517 | 0,0541 | -1,8268 | -0,3878 | -2,1441 | 0,1166 | -2,5840 | -0,2599 | -2,8538 | 1,3239 | -2,5845 | 1,5390 | -1,5846 | 1,5710 | -0,8164 |
| -1,7404 | 1,5295 | -3,9580 | 3,1951 | -2,3317 | 5,2630 | -2,6126 | 7,1439 | -2,2669 | 8,5182 | -2,5845 | 8,6933 | 0,1439 | 7,0202 | 3,1328 | 7,8034 |
| -3,3240 | -3,1035 | -3,9018 | -3,3035 | -3,6927 | -2,2048 | -2,2222 | -1,7138 | -2,9159 | -0,8288 | 1,5390 | 0,1439 | 4,7756 | 1,6371 | 7,7064 | 4,6961 |
| -2,9488 | 0,3099 | -5,1825 | 1,7979 | -3,4815 | 3,9517 | -3,0742 | 5,6271 | -2,7243 | 6,5086 | -1,5846 | 7,0202 | 1,6371 | 6,2028 | 4,9945 | 7,8848 |
| -6,7840 | -5,0428 | -8,5312 | -4,6161 | -7,3924 | -1,8626 | -4,7684 | -0,3435 | -5,4216 | 1,4911 | 1,5710 | 3,1328 | 7,7064 | 4,9945 | 13,4638 | 10,2251 |
| -6,7171 | -2,3516 | -9,7781 | -0,4115 | -7,2608 | 3,1781 | -5,3036 | 5,4821 | -4,6929 | 6,5830 | -0,8164 | 7,8034 | 4,6961 | 7,8848 | 10,2251 | 11,7472 |

$\mathbf{K} = (\mathbf{Y}^{\mathrm{T}}\mathbf{Y} + 1)$

| | | | | | | | | | | | | | | | |
|---|---|---|---|---|---|---|---|---|---|---|---|---|---|---|---|
| 105,2753 | -100,8805 | 126,8302 | -112,9870 | 137,7638 | -85,0252 | 96,6628 | -72,4656 | 20,0609 | -14,6815 | 33,5631 | -26,3899 | 43,7659 | -37,5264 | 53,8849 | -50,8083 |
| -100,8805 | 100,5458 | -122,0135 | 112,4462 | -132,7486 | 85,2002 | -92,6957 | 73,0044 | -18,3913 | 16,8758 | -31,6602 | 28,3961 | -41,6882 | 39,3372 | -51,6437 | 52,3659 |
| 126,8302 | -122,0135 | 153,8893 | -137,6540 | 167,8372 | -104,9678 | 119,7337 | -90,8924 | 35,0146 | -25,8725 | 52,4089 | -41,1596 | 65,5741 | -55,4636 | 78,7726 | -72,2342 |
| -112,9870 | 112,4462 | -137,6540 | 126,7668 | -150,4004 | 97,3128 | -107,0448 | 84,7216 | -31,4382 | 26,5162 | -47,3578 | 40,5354 | -59,4098 | 53,6208 | -71,5118 | 68,9226 |
| 137,7638 | -132,7486 | 167,8372 | -150,4004 | 183,4740 | -115,5416 | 132,1352 | -100,9264 | 44,9350 | -33,2346 | 64,5370 | -50,5826 | 79,3858 | -66,6768 | 94,3558 | -85,3746 |
| -85,0252 | 85,2002 | -104,9678 | 97,3128 | -115,5416 | 76,4196 | -84,7110 | 68,2840 | -37,4148 | 30,3142 | -50,8222 | 42,3600 | -60,9974 | 53,3304 | -71,3804 | 65,8188 |
| 96,6628 | -92,6957 | 119,7337 | -107,0448 | 132,1352 | -84,7110 | 99,1241 | -76,5852 | 53,1096 | -38,3767 | 68,9579 | -52,7598 | 81,0007 | -65,6972 | 93,3892 | -80,2184 |
| -72,4656 | 73,0044 | -90,8924 | 84,7216 | -100,9264 | 68,2840 | -76,5852 | 62,7744 | -46,4672 | 36,6164 | -59,3828 | 48,4456 | -69,2084 | 58,9664 | -79,3904 | 70,6208 |
| 20,0609 | -18,3913 | 35,0146 | -31,4382 | 44,9350 | -37,4148 | 53,1096 | -46,4672 | 120,1665 | -85,3643 | 134,1995 | -99,6723 | 145,0279 | -110,7984 | 157,2529 | -121,0323 |
| -14,6815 | 16,8758 | -25,8725 | 26,5162 | -33,2346 | 30,3142 | -38,3767 | 36,6164 | -85,3643 | 63,6258 | -95,7422 | 74,1691 | -103,7462 | 82,4052 | -112,7567 | 90,0369 |
| 33,5631 | -31,6602 | 52,4089 | -47,3578 | 64,5370 | -50,8222 | 68,9579 | -59,3828 | 134,1995 | -95,7422 | 151,1370 | -112,7827 | 164,1826 | -126,2596 | 178,7551 | -138,9937 |
| -26,3899 | 28,3961 | -41,1596 | 40,5354 | -50,5826 | 42,3600 | -52,7598 | 48,4456 | -99,6723 | 74,1691 | -112,7827 | 87,3069 | -122,8751 | 97,7496 | -134,1131 | 107,6937 |
| 43,7659 | -41,6882 | 65,5741 | -59,4098 | 79,3858 | -60,9974 | 81,0007 | -69,2084 | 145,0279 | -103,7462 | 164,1826 | -122,8751 | 178,9210 | -138,1460 | 195,2875 | -152,7845 |
| -37,5264 | 39,3372 | -55,4636 | 53,6208 | -66,6768 | 53,3304 | -65,6972 | 58,9664 | -110,7984 | 82,4052 | -126,2596 | 97,7496 | -138,1460 | 110,0960 | -151,2800 | 122,0720 |
| 53,8849 | -51,6437 | 78,7726 | -71,5118 | 94,3558 | -71,3804 | 93,3892 | -79,3904 | 157,2529 | -112,7567 | 178,7551 | -134,1131 | 195,2875 | -151,2800 | 213,5665 | -167,9075 |
| -50,8083 | 52,3659 | -72,2342 | 68,9226 | -85,3746 | 65,8188 | -80,2184 | 70,6208 | -121,0323 | 90,0369 | -138,9937 | 107,6937 | -152,7845 | 122,0720 | -167,9075 | 136,2665 |

$\mathbf{G} = \mathbf{X}^{\mathrm{T}}(\mathbf{X}\mathbf{X}^{\mathrm{T}})^{-}\mathbf{X}$

| | | | | | | | | | | | | | | | |
|---|---|---|---|---|---|---|---|---|---|---|---|---|---|---|---|
| 0,2555 | 0,1759 | 0,2282 | 0,1071 | 0,1601 | 0,0237 | 0,1085 | -0,0230 | 0,0710 | -0,0497 | -0,0120 | -0,0659 | -0,0760 | -0,0786 | -0,1259 | -0,1322 |
| 0,1759 | 0,2404 | 0,1123 | 0,2017 | 0,1044 | 0,1730 | 0,0488 | 0,1543 | 0,0435 | 0,0884 | -0,0395 | 0,0539 | -0,0744 | 0,0079 | -0,0976 | -0,0449 |
| 0,2282 | 0,1123 | 0,2206 | 0,0476 | 0,1451 | -0,0377 | 0,1066 | -0,0844 | 0,0654 | -0,0905 | 0,0010 | -0,0962 | -0,0595 | -0,0934 | -0,1083 | -0,1354 |
| 0,1071 | 0,2017 | 0,0476 | 0,1822 | 0,0609 | 0,1780 | 0,0177 | 0,1729 | 0,0240 | 0,1107 | -0,0386 | 0,0788 | -0,0556 | 0,0335 | -0,0646 | -0,0060 |
| 0,1601 | 0,1044 | 0,1451 | 0,0609 | 0,1006 | 0,0073 | 0,0692 | -0,0226 | 0,0447 | -0,0371 | -0,0060 | -0,0461 | -0,0465 | -0,0522 | -0,0783 | -0,0850 |
| 0,0237 | 0,1730 | -0,0377 | 0,1780 | 0,0073 | 0,2080 | -0,0239 | 0,2214 | -0,0004 | 0,1565 | -0,0422 | 0,1242 | -0,0361 | 0,0741 | -0,0261 | 0,0483 |
| 0,1085 | 0,0488 | 0,1066 | 0,0177 | 0,0692 | -0,0239 | 0,0517 | -0,0467 | 0,0313 | -0,0477 | 0,0017 | -0,0495 | -0,0274 | -0,0468 | -0,0511 | -0,0661 |
| -0,0230 | 0,1543 | -0,0844 | 0,1729 | -0,0226 | 0,2214 | -0,0467 | 0,2447 | -0,0140 | 0,1794 | -0,0435 | 0,1475 | -0,0248 | 0,0954 | -0,0043 | 0,0776 |
| 0,0710 | 0,0435 | 0,0654 | 0,0240 | 0,0447 | -0,0004 | 0,0313 | -0,0140 | 0,0199 | -0,0193 | -0,0019 | -0,0227 | -0,0201 | -0,0246 | -0,0345 | -0,0388 |
| -0,0497 | 0,0884 | -0,0905 | 0,1107 | -0,0371 | 0,1565 | -0,0477 | 0,1794 | -0,0193 | 0,1357 | -0,0298 | 0,1148 | -0,0080 | 0,0790 | 0,0132 | 0,0731 |
| -0,0120 | -0,0395 | 0,0010 | -0,0386 | -0,0060 | -0,0422 | 0,0017 | -0,0435 | -0,0019 | -0,0298 | 0,0088 | -0,0229 | 0,0094 | -0,0126 | 0,0088 | -0,0059 |
| -0,0659 | 0,0539 | -0,0962 | 0,0788 | -0,0461 | 0,1242 | -0,0495 | 0,1475 | -0,0227 | 0,1148 | -0,0229 | 0,0996 | 0,0012 | 0,0718 | 0,0234 | 0,0725 |
| -0,0760 | -0,0744 | -0,0595 | -0,0556 | -0,0465 | -0,0361 | -0,0274 | -0,0248 | -0,0201 | -0,0080 | 0,0094 | 0,0012 | 0,0267 | 0,0119 | 0,0395 | 0,0307 |
| -0,0786 | 0,0079 | -0,0934 | 0,0335 | -0,0522 | 0,0741 | -0,0468 | 0,0954 | -0,0246 | 0,0790 | -0,0126 | 0,0718 | 0,0119 | 0,0563 | 0,0330 | 0,0646 |
| -0,1259 | -0,0976 | -0,1083 | -0,0646 | -0,0783 | -0,0261 | -0,0511 | -0,0043 | -0,0345 | 0,0132 | 0,0088 | 0,0234 | 0,0395 | 0,0330 | 0,0630 | 0,0609 |
| -0,1322 | -0,0449 | -0,1354 | -0,0060 | -0,0850 | 0,0483 | -0,0661 | 0,0776 | -0,0388 | 0,0731 | -0,0059 | 0,0725 | 0,0307 | 0,0646 | 0,0609 | 0,0862 |

Appendix B. Algorithm implementation within the Octave language.

```
function [R, alfa, S, invXX, W, w0, b, H, K, G] = prepareSHM
(d, X, Y)
invXX=inv(X*X');
G=X'*invXX*X;
K=(Y'*Y+ 1);
H=(d*ones(1,length(X))).*K.*G.*(ones(length(X),1)*d');
alfa =  pqpnonneg (H, (-ones(size(d))));
sum=zeros(size(invXX));
sum0=zeros(length(invXX),1);
for i=1:length(X)
    sum=sum+alfa(i)*d(i)*X(:,i)*(Y(:,i))';
    sum0=sum0+alfa(i)*d(i)*X(:,i);
end
W=invXX*sum;
w0=invXX*sum0;
I=find((sign((fix (alfa*10000))/10000)).*sign(d+1) > 0);
I=I(1,1);
b=-(X(:,I))'*W*Y(:,I) - w0'*X(:,I) + 1;
W1=[w0,W];
Y1=[ones(1,length(X));Y];
R=diag(X'*W1*Y1) + b;
S=[X'*W,X'*w0+b-d];
S=S((fix (alfa*10000))/10000 ~= 0,:);
```

Table B.1 Correspondence between the script variables and data objects used within the text and in Appendix A.

| | |
|---|---|
| R | $[h(\mathbf{x}_i, \mathbf{y}_i), \ \cdots]$ |
| alfa | $\boldsymbol{\alpha}$ |
| S | $[\mathbf{X}^{\mathrm{T}}\widehat{\mathbf{W}} \quad \mathbf{X}^{\mathrm{T}}\widehat{\mathbf{w}}_0 + \mathbf{e}b - \mathbf{d}]$ |
| invXX | $(\mathbf{X}\mathbf{X}^{\mathrm{T}})^{-}$ |
| W | $\widehat{\mathbf{W}}$ |
| w0 | $\widehat{\mathbf{w}}_0$ |
| b | $b$ |
| H | $\mathbf{H}$ |
| K | $\mathbf{K}$ |
| G | $\mathbf{G}$ |
| d | $\mathbf{d}$ |
| X | $\mathbf{X}$ |
| Y | $\mathbf{Y}$ |

## References.